%% file: main.tex
\documentclass[preprint, 12pt, a4paper]{elsarticle}

\usepackage{amssymb}

\usepackage{float}
\restylefloat{table}

\usepackage[hidelinks]{hyperref} 
\usepackage[T1]{fontenc} 

\newcommand{\inlinecode}[1]{\texttt{#1}} 
\newcommand{\frameworkname}{\textit{schlably}} 
\newcommand{\inlineskip}{\vskip 5mm}

\usepackage{xcolor} 

\usepackage{array} 
\usepackage{multirow} 
\usepackage{makecell} 
\usepackage{rotating} 
\usepackage{tabularx} 
\usepackage[nointegrals]{wasysym} 

\journal{SoftwareX}

\begin{document}

\begin{frontmatter}



\title{\frameworkname{}: A Python Framework for Deep Reinforcement Learning Based Scheduling Experiments}


\author{Constantin Waubert~de~Puiseau, Jannik Peters, Christian D\"orpelkus, Hasan Tercan, Tobias Meisen}

\address{
    Institute for Technologies and Management of the Digital Transformation
    \linebreak
    University of Wuppertal
    \linebreak
    Rainer-Gruenter-Str.~21, Wuppertal, 42119, NRW, Germany
}

\begin{abstract}
Research on deep reinforcement learning (DRL) based production scheduling (PS) has gained a lot of attention in recent years, primarily due to the high demand for optimizing scheduling problems in diverse industry settings. Numerous studies are carried out and published as stand-alone experiments that often vary only slightly with respect to problem setups and solution approaches. The programmatic core of these experiments is typically very similar. Despite this fact, no standardized and resilient framework for experimentation on PS problems with DRL algorithms could be established so far. In this paper, we introduce \frameworkname{}, a Python-based framework that provides researchers a comprehensive toolset to facilitate the development of PS solution strategies based on DRL. \frameworkname{} eliminates the redundant overhead work that the creation of a sturdy and flexible backbone requires and increases the comparability and reusability of conducted research work.
\end{abstract}


\end{frontmatter}

\section*{Required Metadata}
\label{sec:required-metadata}

\section*{Current code version}
\label{sec:current-code-version}

\begin{table}[H]
\begin{tabular}{|l|p{6.5cm}|p{6.5cm}|}
\hline
\textbf{Nr.} & \textbf{Code metadata description} & \textbf{Please fill in this column} \\
\hline
C1 & Current code version & v0.1.0 \\
\hline
C2 & Permanent link to code/repository used for this code version & \url{https://github.com/tmdt-buw/schlably} \\
\hline
C4 & Legal Code License   & Apache License, 2.0 (Apache-2.0) \\
\hline
C5 & Code versioning system used & git \\
\hline
C6 & Software code languages, tools, and services used & Python, OpenAI Gym, RLlib, Weights and Biases \\
\hline
C7 & Compilation requirements, operating environments \& dependencies & Python 3.10\\
\hline
C8 & If available Link to developer documentation/manual & \url{https://schlably.readthedocs.io/en/latest/index.html} \\
\hline 
C9 & Support email for questions & schlably@uni-wuppertal.de\\
\hline
\end{tabular}
\caption{Code metadata (mandatory)}
\label{tab:code-metadata} 
\end{table}

\section{Motivation and Significance}
\label{sec:motivation}
Production scheduling (PS) is a challenging and highly researched problem in operations research (OR) and optimization. It deals with allocating resources to production jobs over time to minimize criteria such as time, effort, and cost \cite{Pinedo.2016}. PS problems belong to the class of combinatorial optimization problems which require much computation effort to be solved sufficiently well with increasing problem sizes. In recent years, with increasingly powerful algorithms and computational hardware, deep reinforcement learning (DRL) has emerged as a promising tool to address PS problems \cite{bello.2016, zhang.2020, kuhnle.2020, vanEkeris.2021, samsonov., Puiseau.2022, SebastianPol.2021}. DRL is a machine learning paradigm, in which deep learning models are trained through experience data gathered by a reinforcement learning agent that interacts with its environment, thereby autonomously deriving solution strategies for sequential tasks \cite{Sutton.2018}.

The young research field of DRL-based PS lies at the intersection of operations research and artificial intelligence and as such is characterized by a heterogeneous community with varying problem-solving approaches and technical skill sets. Yet, all empirical studies apply a very similar experimental setup consisting of the same main software components: An environment representing the production facility layout and logic, a scheduling problem generator, a DRL agent algorithm, and logging as well as evaluation tools. The difference between different experimental setups usually lies within one or more of these components, for example by incorporating a new DRL algorithm \cite{rinciog., monaci.2021, Luo.2020}, interaction logic between agent and environment \cite{samsonov., zhang.2020}, training procedure \cite{Iklassov.09.06.2022b}, learning objective \cite{sakr.2021, Luo.2020} or additional problem constraint \cite{sakr.2021, luo.2021}. Regardless of large overlaps, all researchers implement their own individual experimentation framework with the following two consequences: Large initial ramp-up efforts when experimenting with new methodologies or custom problem settings, and scarcity of empirical comparisons to other works.
In this paper, we address these shortcomings and present the software framework \frameworkname{} for developing and evaluating DRL-based PS solutions. \frameworkname{} provides the following contributions: 
\begin{itemize}
    \item It is modular, so individual changes may be adapted without much overhead.
    \item It works out of the box with functioning environments, data-generation scripts, agents, logging functions, training, and testing scripts.
    \item It provides benchmark datasets for different scheduling problem classes and sizes.
    \item It includes widely recognized benchmark algorithms and results.
    \item It facilitates the application of algorithms designed for one problem class and size to other problems.
\end{itemize}

\frameworkname{} will accelerate the research area of DRL-based PS under real-world conditions by lowering the barrier of entry for researchers from different domains with different perspectives, levels of expertise, and objectives.

\section{Background and Related Work}
\label{sec:background}
\frameworkname{} started as a code base for experiments on a real-world inspired scheduling problem in the context of a university research project with industrial partners. As such, several requirements became apparent early on that can be summarized in four general \textbf{design goals}. First, from the applied industrial perspective, \frameworkname{} has to offer the integration of DRL methods and heuristics which work out of the box. Second, it also has to cover different scheduling scenarios, e.g. including such bounded by resource constraints. Third, from the scientific research perspective, \frameworkname{} should support detailed comparable evaluations of methods. Lastly, it has to be easy to interact with at the code level, to enable students with limited experience to quickly understand the topic, concepts, and implementation. The implications of these design goals are discussed in this section.

In the following, we present an overview of related published experimental frameworks and compare them to our design goals in \frameworkname{}. In the comparison, we include frameworks dedicated to being used by others \cite{Tassel.2021, Zheng.2022, DrilyassPHx.2022, Tassel.2022, Samsonov.2022} and frameworks published in a supplementary fashion along with research papers and projects \cite{tejaswinimedi.2022, Venturelli.2015, samybarrech.2018, Zhang.2020c, Kumar.2019, vanEkeris.2020}, because in practice both may serve as starting points for additional experiment designs. The frameworks were found via references in scientific publications and a search of "Scheduling Reinforcement Learning" on GitHub. We do not claim the list to be exhaustive but are not aware of any other popular frameworks at the time of writing this paper. Commercial or proprietary scheduling software was excluded because license fees and other accompanying challenges introduce a major barrier. However, we are not aware of any commercial software that provides the tight integration of DRL and PS. Table \ref{tab:1} provides an overview of the frameworks. In addition, we assessed them regarding their fulfillment of our design goals which are formally categorized into the following four groups.

\input{framework_comparison_table}%

\inlineskip
\noindent
\paragraph{\textbf{Pre-Implemented Benchmarks}}
Several frameworks either provide pre-implemented DRL agents and scripts for training or the easy integration of agents from popular DRL libraries, e.g. from StableBaselines \cite{stable-baselines3} or RLlib \cite{EricLiang.2018}. Like \cite{Samsonov.2022}, our goal is to enable and facilitate both, the manual extensions of basic DRL algorithms, like Deep Q-Networks (DQN) \cite{mnih.2013} and Proximal Policy Optimization (PPO) \cite{schulman.2017}, as well as the usage of powerful third-party libraries. Both options are important to empower users to choose the appropriate approach for their respective research interests and are therefore implemented in \frameworkname{}. 
Additionally, for the sake of comparability, it is crucial to provide common benchmarks in the form of popular priority dispatching rules (PDRs), such as Shortest-Processing-Time-First, and a flexible optimal solver that can handle several scheduling problem types. \frameworkname{} provides these baselines, while most other frameworks only cover a few PDRs, often missing competitive ones, such as Most-Tasks-Remaining and even random baselines, where agent actions are sampled from a normal distribution.

\inlineskip
\noindent
\paragraph{\textbf{Scheduling Instance Generation}}
Generating new data with varying problem cases is necessary to enable comprehensive training and testing of a DRL agent. Accordingly, a suitable framework must implement a flexible problem instance generator. This generator enables the user to create scheduling problems of different popular categories (e.g. the Job Shop Scheduling Problem (JSSP) or the Flexible Job Shop Scheduling Problem (FJSSP) \cite{Pinedo.2016}) instances with any combination of instance variables, like the number of jobs, number of tasks, runtimes, and more. All combinations and values are drawn using equal distributions. \frameworkname{} further extends these options by an optional resource constraint, in which each task requires a certain additional resource. These options are already implemented in the data generation and can be processed by the agents provided in schlably. Moreover, its design simplifies the integration of additional scheduling problem types to encourage the implementation of individual, more complex, or more specific use cases.
Finally, to the best of our knowledge, this is the first framework providing an optional resource constraint. Concretely, the user is able to specify a required tool per operation.

\inlineskip
\noindent
\paragraph{\textbf{Logging and Evaluation}}
Logging results and evaluation metrics in a structured manner is key for quick feedback during training runs, but also for identifying patterns in and drawing conclusions from large-scale experiments. Our objective with \frameworkname{} is to provide extensive logging options that may be turned on and off, and where results and models may be shared to promote collaboration on projects. This design goal is met to a large degree by \cite{Samsonov.2022} from which we took much inspiration in this regard. All other frameworks do not address this design goal.
For the evaluation of solutions generated by DRL agents, a comprehensive framework should apply the benchmarks mentioned above and provide an overview of the overall performance. Most of the reviewed frameworks lack functionality in this aspect. 
Moreover, to get a graphical overview and to visually support the tracking of very specific actions in a production schedule, a Gantt chart plotter is useful for human inspection. The Gantt chart should display all metadata of operations, e.g. the runtime or required tool, and has been found to be helpful for debugging and evaluating DRL agents. Many other frameworks, but not all, include a Gantt chart plotter.
 
\inlineskip
\noindent
\paragraph{\textbf{Code Usability}}
As usability is of utmost importance, a framework has to offer easy access through full documentation and must include a README, user application programming interface (API) manual, and formal functionality description. Within the reviewed frameworks, only \cite{Samsonov.2022} covers all criteria. To be usable with different skill sets our explicit goal is to enable users to start experimenting with small design decisions by only using the configuration files but at the same time facilitate substantial logical changes to routines and components by means of their own implementations. For that reason, we would favor comprehensibility over efficiency wherever a trade-off is unavoidable. This requires a careful balance. In our opinion, all other frameworks overemphasize one side: \cite{Samsonov.2022} offers many functional changes through configuration files but at the cost of a comparatively complex software architecture. On the other hand, all other frameworks are much smaller and easier to get an overview of, at the expense of limited functionality. Lastly, a framework with a claim to widespread use should stick to conventional APIs. In the context of DRL, the most commonly used API is the OpenAI Gym \cite{1606.01540} API. Only half of the reviewed frameworks adhere to it.

\section{Software Architecture}
\label{sec:software-architecture}

\begin{figure}[htbp]
  \centering
  \includegraphics[width=\textwidth]{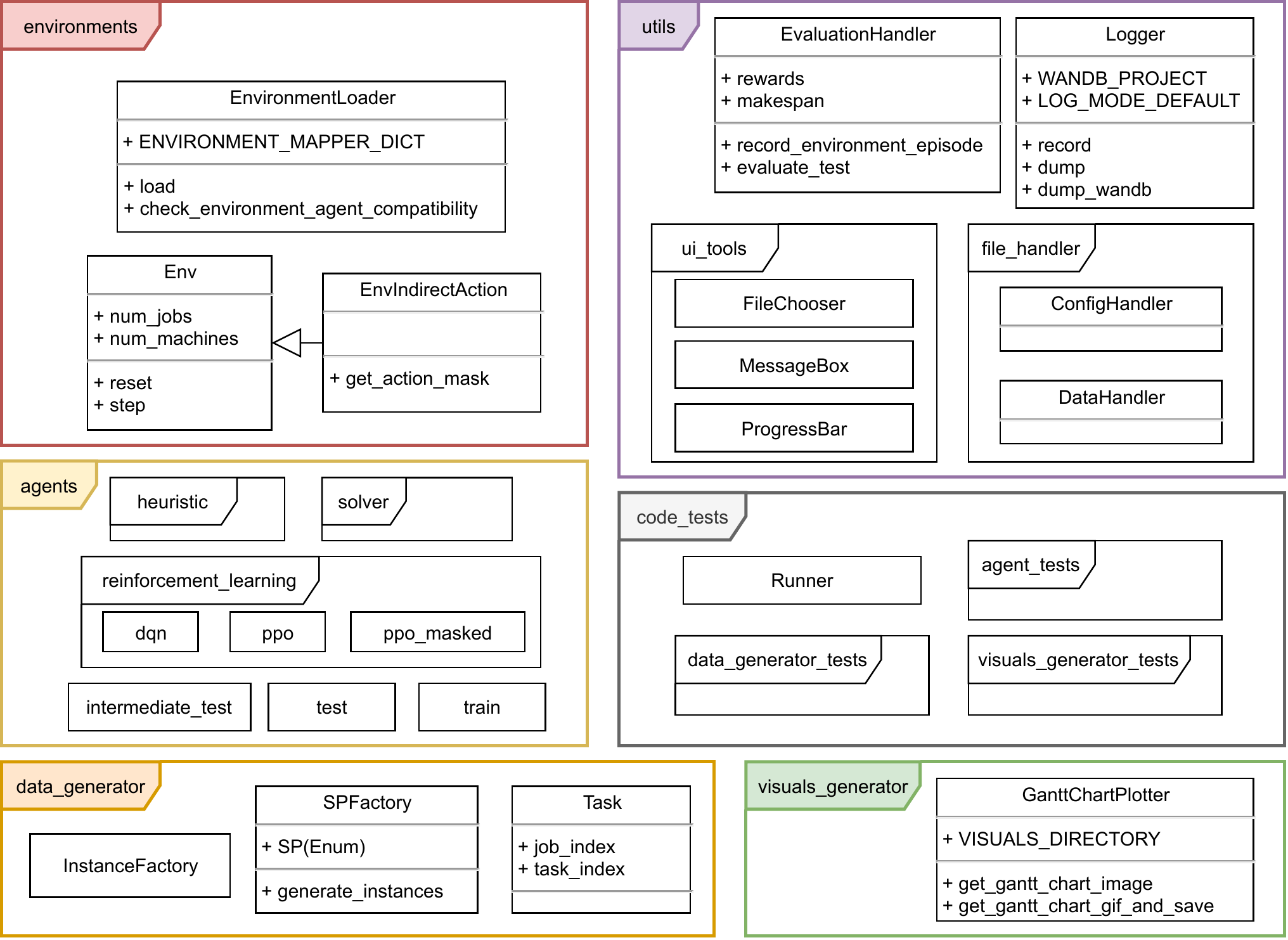}
  \caption[]{Overview of \frameworkname{} project and code structure.}
  \label{fig:architecture_src_overview}
\end{figure}

This chapter describes our framework with focus on implementation-specific details. We are providing a general overview of the code itself, focusing on currently existing exemplary implementations while also pointing out open interfaces. Furthermore, we describe details regarding the main components of \frameworkname{} to demonstrate the realization of the design goals, as introduced in Chapter~\ref{sec:background}, and to enable users to fit \frameworkname{} to their needs. The overall structure is illustrated in Figure \ref{fig:architecture_src_overview}. We divided the code base into six main components, which are described below in detail. Following this component-oriented approach, and in combination with comprehensive code documentation, \frameworkname{} adheres to the objective of the fourth design goal, which requires easy interaction and usability at the code level.

\inlineskip
\noindent
\paragraph{\textbf{Data generator}}
The general data structure of scheduling problems, as used in \frameworkname{}, is represented by so-called instances. A user can generate infinite instances of a scheduling problem, however, each instance is a specific configuration and entity. The specific configuration, contained within an instance, is given by a number of jobs, with a job simply being an encompassing logical container consisting of individual tasks.
The \inlinecode{data\_generator} component incorporates the necessary classes to generate such an instance and the individual tasks. From the scheduling problem point-of-view, it is the centerpiece of the problem formulation and representation. The \inlinecode{Task} class is a specifically designed data class and its entities are the atomic units of a scheduling problem instance. Such an instance can be created via the \inlinecode{SPFactory}, which allows the generation of different types of scheduling problems that are given via the included \inlinecode{Enum}. If users would like to introduce a new type of scheduling problem into \frameworkname{}, they would have to include their function in this class and add it to the \inlinecode{Enum}. Finally, the \inlinecode{InstanceFactory} enables high-level access to the problem factory class and manages the configuration-based creation of batches of instances. Thus, the \inlinecode{data\_generator} component realizes the foundation of the second design goal, which requires the implementation and handling of different scheduling scenarios. 

\inlineskip
\noindent
\paragraph{\textbf{Environment}}
An environment defines the observation space, action space, and reward strategy. Thus, it represents a simulation of the agent's environment and interaction dynamics and is the central piece of any DRL approach.
All \frameworkname{} environments are included in the \inlinecode{environments} component. Exemplary, we provide a simple scheduling \inlinecode{Env} as well as a derived version named \inlinecode{EnvIndirectAction} to showcase the expandability. All environments adhere to the Gym API and are explicitly derived from a base Gym environment. The \inlinecode{EnvironmentLoader} class enables high-level access and management of the different environment types and appropriate algorithms, as not all algorithms are feasible for every environment. New environments have to be included in this component and added to the managing \inlinecode{EnvironmentLoader}. This encapsulated approach, in conjunction with the \inlinecode{data\_generator} component, represents the implementation of the second design goal.

\inlineskip
\noindent
\paragraph{\textbf{Agent}}
The \inlinecode{agents} component combines the heuristic functionalities, the solver, and implementations of DRL algorithms as well as the \inlinecode{train} and \inlinecode{test} functions for the DRL approach. Users can to integrate functionalities from other DRL frameworks, like more extensive training procedures, model types, and learning algorithms via pre-defined interfaces. As such, the \inlinecode{agents} component realizes the first design goal, to support and simplify the integration of out-of-the-box-methods as well as pre-implemented benchmarks.

\inlineskip
\noindent
\paragraph{\textbf{Visual generator}}
The component \inlinecode{visuals\_generator} incorporates all classes and scripts which are used to create visualizations of the problem instances and generated solutions. These functionalities are intentionally isolated as different scheduling problem environments and still share the same visualization approach. \frameworkname{}, for example, introduces a \inlinecode{GanttChartPlotter} that enables a user to generate individual Gantt chart images (see Figure~\ref{fig:agent_diff_pic}b) or create a GIF of the scheduling progress. Thus, it is part of the implementation of the third design goal. The module is intended for debugging and visual analysis and can currently only display Gantt-Charts of problem sizes smaller than 8x8 because of limitations of the used library (the exact limit depends on the number of tasks and processing times). However, we believe that visualizations become less useful for larger problem sizes, because there are too many blocks and colors to gain an overview.

\inlineskip
\noindent
\paragraph{\textbf{Utils}}
The \inlinecode{utils} component aggregates classes and functions which have a supporting character for the main functionalities of \frameworkname{}. Specifically, it includes user interface components (\inlinecode{ui\_tools}), data interface components (\inlinecode{file\_handler}), e.g. to load and save data, and the high-level \inlinecode{Logger} class. Accordingly, the \inlinecode{utils} component realizes the third design goal, facilitating logging and evaluations for comparisons.

\inlineskip
\noindent
\paragraph{\textbf{Code tests}}
All code tests that ensure the crucial functionality of the described components are collected in the \inlinecode{code\_tests} component. Up to this point, we included multiple unit tests with a central \inlinecode{Runner}. These are also intended as an example for users that plan to extend the code base.

\section{Illustrative Example}
\label{sec:illustrative-example}
\begin{figure}
  \centering
  \includegraphics[width=\textwidth]{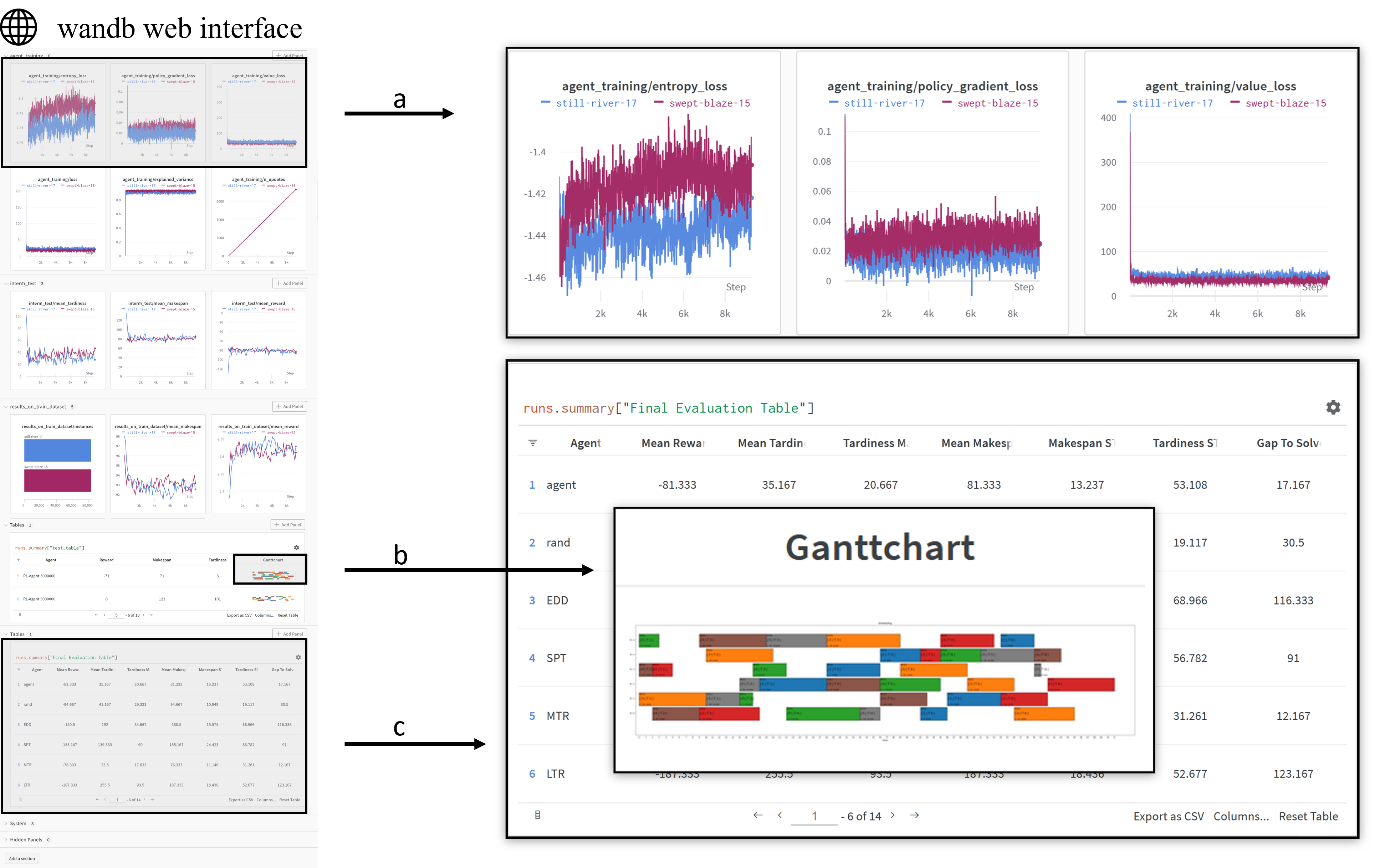}
  \caption[]{Comparing agent runs in Weights\&Biases (screenshot from the web interface shown on the left-hand side). a) Visualized training curves for interpreting the learning performance of the agent. b) Gantt chart depicting the solution of the trained agent on a selected test instance. c) Table providing evaluation results and comparison of the trained agents and benchmark methods on the test instances.}
  \label{fig:agent_diff_pic}
\end{figure}

To illustrate a typical use case, we consider a scenario in which an ML engineer wants to compare the learning behavior of two PPO agents that interact with the implemented environment. It is also part of our tutorial in the documentation. One agent is trained on 6x6 JSSP instances and receives a reward based on the change in the time to complete all tasks (i.e. the makespan) per step, as proposed in \cite{zhang.2020}. This setup is also the default setting delivered in the framework. The other one is trained on a 3x4 tool-constrained JSSP instance and receives a zero reward per step with the exception of the last step, where the reward is equal to the overall achieved makespan. The way in which both agent interacts with the default environment is that in each step, the agent chooses between the next unscheduled task within each job sequence. This task is then integrated into the current schedule by scheduling it at the earliest time possible in accordance with the constraints and without shifting already scheduled tasks. The remaining training parameters are kept constant. The second training requires only minimal manual changes to the base model. These include setting different configuration parameters, generating new data, and changing the reward function in the base environment. Details may be found in the documentation.
The integrated interface to Weights\&Biases \cite{Biewald.2020} makes it easy to compare the training curves and achieved results, as depicted in Figure \ref{fig:agent_diff_pic}. 

The described short example reflects several of our design goals. Figure \ref{fig:agent_diff_pic}c) demonstrates that the agents' performance is automatically compared to many other benchmarks and with respect to different dimensions such as the reward or the gap to the optimal solver. The continuous logging and graphical depiction are visible in Figure \ref{fig:agent_diff_pic}a) and b). The example also showcases our understanding of high code usability. The experiments could be defined by changing training parameters (only a few lines in configuration files) and minimal intended changes to the source code. Examples of the most common changes which are intended to be coded are explained in more detailed follow-along tutorials in the provided documentation.

\section{Impact}
\label{sec:impact}
\frameworkname{} is useful for the entire community around PS with DRL. Compared to other frameworks, it is particularly useful to reduce the entry barrier for researchers from the OR or other related domains, who want to empirically explore a new methodology for scheduling problems, and for DRL researchers who want to test a new algorithm on a challenging and impactful problem domain. We believe that the seamless interchangeability of problem settings offered by \frameworkname{} will also encourage researchers in the domain of PS with DRL to try out methodologies applied to one particular problem setting (e.g. 6x6 JSSP) on different problem settings (e.g. 11x11 tool-constrained JSSP). This has the potential to greatly speed up the transfer of research from academic problems to real-world problems. 

In several projects where our test partners and we have used \frameworkname{}, it has significantly increased the throughput of experiments. This is achieved because new methodological ideas can be integrated more quickly and the results of experiments can be compared more easily. \frameworkname{} facilitates the generation of new problem instances and the training and evaluation of custom DRL agents. Due to the various pre-implementations in the framework, such as training and testing routines, well-known scheduling benchmarks, and visualization of logged results, it is much easier to conduct experimental research in DRL for PS. In addition, collaboration has become more effective because design changes can be compared easily and the results of peers can be viewed online through Weights\&Biases. We have further experienced a substantial increase in productivity in research projects, where new researchers and university students, who had no prior domain knowledge and little coding skills, had to conduct experiments on the PS domain. This, we mainly attribute to the code documentation and modular structure, but also to the fact that \frameworkname{} is 100\% written in Python and therefore runs on all relevant operation systems.

\section{Discussion and limitations}
\label{sec:discussion_and_limitations}

In its current state, \frameworkname{} serves as a useful framework for empirical DRL-based PS research. It has reached a maturity level, at which it works out-of-the-box and, to the best of our knowledge, offers the broadest range of different easy-to-implement design choices compared to any published framework. \frameworkname{}, on the one hand, is intended to be abstract and modular enough to offer different instance generation, training, and testing configurations without many lines of code. On the other hand, it is designed to not be too interwoven in its code structure to hinder the extension with fundamentally different features experts might find desirable. As such, the development required a balancing act and certain compromises, which some may see as limitations.
For example, one deliberate choice was made in favor of a class-based problem description as opposed to a vector representation. The class-based description simplifies the search and usage of certain information about the current state of jobs and increases code readability compared to a vector problem representation. Hence, the choice was made between readability and computational efficiency in favor of the former.

\section{Conclusions}
\label{sec:conclusions}
In this paper, we introduced \frameworkname{}, a software framework for research on DRL-based PS. With the release of the framework, we strive towards two main goals:
the first is to lower the entry barrier for researchers, who have little experience with production scheduling, deep reinforcement learning (DRL) and/or coding. The second goal is to encourage researchers already active in the field to apply and test their methods on other problem settings, which is largely facilitated by \frameworkname{}. Both goals aim at promoting the transfer of DRL methods to real-world scheduling applications.
In the future, we plan to include more problem settings, such as the dynamic JSSP and stochastic properties of environments like machine breakdowns to get even closer to real-world scenarios.

\section{Conflict of Interest}
We wish to confirm that there are no known conflicts of interest associated with this publication and there has been no significant financial support for this work that could have influenced its outcome.

\section*{Acknowledgements}
\label{sec:acknowledgements}
This research work was undertaken within the research project AlphaMES funded by the German Federal Ministry for Economic Affairs and Climate Action (BMWK).



\bibliographystyle{elsarticle-num} 
\bibliography{bibliography.bib}




\section*{Current executable software version}
\label{sec:current-executable-software-version}

Ancillary data table required for sub version of the executable software: (x.1, x.2 etc.) kindly replace examples in right column with the correct information about your executables, and leave the left column as it is.

\begin{table}[!hp]
\begin{tabular}{|l|p{6.5cm}|p{6.5cm}|}
\hline
\textbf{Nr.} & \textbf{(Executable) software metadata description} & \textbf{Please fill in this column} \\
\hline
S1 & Current software version & v0.1.0 \\
\hline
S2 & Permanent link to executables of this version  & \url{https://github.com/tmdt-buw/schlably} \\
\hline
S3 & Legal Software License & Apache License, 2.0 (Apache-2.0) \\
\hline
S4 & Computing platforms/Operating Systems & Python, OpenAI Gym, DRLlib,Weights and Biases \\
\hline
S5 & Installation requirements \& dependencies &  Python 3.10 \\
\hline
S6 & If available, link to user manual - if formally published include a reference to the publication in the reference list & \url{https://schlably.readthedocs.io/en/latest/index.html} \\
\hline
S7 & Support email for questions & schlably@uni-wuppertal.de \\
\hline
\end{tabular}
\caption{Software metadata (optional)}
\label{tab:software-metadata} 
\end{table}

\end{document}

%% file: framework_comparison_table.tex
\begin{table}
\centering
\begin{tabularx}{\textwidth}{l>{\tiny}l*{12}{>{\hspace{-4pt}}c<{\hspace{-4pt}}}}
                                                												  &                                               & \cite{Tassel.2021} 	 & \cite{Zheng.2022}   & \cite{DrilyassPHx.2022}   & \cite{tejaswinimedi.2022}   & \cite{Tassel.2022}   & \cite{Samsonov.2022}   & \cite{Venturelli.2015}   & \cite{samybarrech.2018}   & \cite{Zhang.2020c}   & \cite{Kumar.2019}   & \cite{vanEkeris.2020}   & schlably   \\
\hline
\multirow{4}{*}{\textbf{B}}     & Implemented RL-agents                     & \RIGHTcircle         & \Circle             & \RIGHTcircle              & \RIGHTcircle                 & \Circle              & \CIRCLE                & \Circle                  & \Circle                   & \RIGHTcircle         & \Circle             & \RIGHTcircle            & \CIRCLE  \\
								& Implemented PDRs            & \Circle              & \CIRCLE             & \RIGHTcircle              & \Circle                      & \Circle              & \Circle                & \Circle                  & \CIRCLE                   & \Circle              & \Circle             & \CIRCLE                 & \CIRCLE  \\
								& Implemented opt. solver                & \Circle              & \CIRCLE             & \Circle                   & \Circle                      & \Circle              & \CIRCLE                & \CIRCLE                  & \Circle                   & \Circle              & \Circle             & \CIRCLE                 & \CIRCLE  \\
								& Interface for RLLib                           & \Circle              & \Circle             & \CIRCLE                   & \Circle                      & \RIGHTcircle         & \Circle                & \Circle                  & \Circle                   & \Circle              & \Circle             & \Circle                 & \CIRCLE  \\
\hline
\multirow{5}{*}{\textbf{S}}		& Flexible Data Generation                               & \Circle              & \Circle             & \Circle                   & \Circle                      & \Circle              & \CIRCLE                & \Circle                  & \Circle                   & \CIRCLE              & \RIGHTcircle        & \CIRCLE                 & \CIRCLE  \\
								& JSSP                                          & \CIRCLE              & \CIRCLE             & \CIRCLE                   & \Circle                      & \CIRCLE              & \CIRCLE                & \CIRCLE                  & \CIRCLE                   & \CIRCLE              & \CIRCLE             & \CIRCLE                 & \CIRCLE  \\
								& FJSSP                                         & \Circle              & \Circle             & \Circle                   & \CIRCLE                      & \Circle              & \Circle                & \Circle                  & \CIRCLE                   & \Circle              & \Circle             & \Circle                 & \CIRCLE  \\
                                & Different problem types                       & \Circle              & \CIRCLE             & \RIGHTcircle              & \CIRCLE                      & \CIRCLE              & \CIRCLE                & \RIGHTcircle             & \CIRCLE                   & \CIRCLE              & \CIRCLE             & \CIRCLE                 & \CIRCLE  \\
                                & Resource constraint tool                     & \Circle              & \Circle             & \Circle                   & \Circle                      & \Circle              & \Circle                & \Circle                  & \Circle                   & \Circle              & \Circle             & \Circle                 & \CIRCLE  \\
\hline
\multirow{5}{*}{\textbf{L}}     & Log achieved results                          & \Circle              & \RIGHTcircle        & \RIGHTcircle              & \RIGHTcircle                 & \Circle              & \CIRCLE                & \Circle                  & \Circle                   & \RIGHTcircle         & \Circle             & \RIGHTcircle            & \CIRCLE  \\
                                & Evaluate achieved results                     & \Circle              & \Circle             & \Circle                   & \RIGHTcircle                 & \Circle              & \CIRCLE                & \Circle                  & \RIGHTcircle              & \RIGHTcircle         & \RIGHTcircle        & \CIRCLE                 & \CIRCLE  \\
                                & Visualize Gantt-Chart                    & \Circle              & \CIRCLE             & \Circle                   & \CIRCLE                      & \CIRCLE              & \CIRCLE                & \Circle                  & \CIRCLE                   & \CIRCLE              & \RIGHTcircle        & \Circle                 & \CIRCLE  \\
                                & Comparison to solver                          & \Circle              & \Circle             & \Circle                   & \Circle                      & \Circle              & \CIRCLE                & \Circle                  & \Circle                   & \Circle              & \Circle             & \CIRCLE                 & \CIRCLE  \\
                                & Comparison to PDRs      & \Circle              & \Circle             & \RIGHTcircle              & \Circle                      & \Circle              & \Circle                & \Circle                  & \Circle                   & \Circle              & \Circle             & \CIRCLE                 & \CIRCLE  \\
\hline
\multirow{7}{*}{\textbf{C}}     & Paper                                         & \Circle              & \CIRCLE             & \Circle                   & \Circle                      & \CIRCLE              & \CIRCLE                & \Circle                  & \Circle                   & \CIRCLE              & \Circle             & \CIRCLE                 & \CIRCLE  \\
                                & README                                        & \RIGHTcircle         & \CIRCLE             & \CIRCLE                   & \CIRCLE                      & \CIRCLE              & \CIRCLE                & \CIRCLE                  & \CIRCLE                   & \Circle              & \CIRCLE             & \CIRCLE                 & \CIRCLE  \\
                                & Code documentation                            & \Circle              & \CIRCLE             & \Circle                   & \Circle                      & \Circle              & \Circle                & \CIRCLE                  & \CIRCLE                   & \Circle              & \CIRCLE             & \RIGHTcircle            & \CIRCLE  \\
                                & Easily personalizable   & \Circle              & \Circle             & \CIRCLE                   & \Circle                      & \Circle              & \Circle                & \Circle                  & \Circle                   & \Circle              & \Circle             & \Circle                 & \CIRCLE  \\
                                & Works out-of-the-box & \CIRCLE              & \RIGHTcircle        & \CIRCLE                   & \Circle                      & \Circle              & \CIRCLE                & \CIRCLE                  & \CIRCLE                   & \CIRCLE              & \Circle             & \CIRCLE                 & \CIRCLE  \\
                                & User manual in Readme      & \CIRCLE              & \RIGHTcircle        & \RIGHTcircle              & \Circle                      & \RIGHTcircle         & \CIRCLE                & \RIGHTcircle             & \RIGHTcircle              & \RIGHTcircle         & \CIRCLE             & \Circle                 & \CIRCLE  \\
                                & OpenAI Gym Env                                       & \Circle              & \CIRCLE             & \Circle                   & \Circle                      & \CIRCLE              & \CIRCLE                & \Circle                  & \Circle                   & \CIRCLE              & \CIRCLE             & \CIRCLE                 & \CIRCLE \\ \rule{0pt}{4ex}
%
%
\end{tabularx}
\begin{tabular}{c}
    Legend: \hspace{0.2cm} \Circle not fulfilled \hspace{0.2cm} \RIGHTcircle half-fulfilled \hspace{0.2cm} \CIRCLE fulfilled \\
\end{tabular}
	\caption{Overview of related frameworks and their fulfillment regarding %
	pre-implemented benchmarks (\textbf{B}), %
	scheduling instances (\textbf{S}), %
	logging and evaluation (\textbf{L}), %
	and code usability (\textbf{C})}
	\label{tab:1}
\end{table}